\documentclass[final]{cvpr}

\usepackage{times}
\usepackage{epsfig}
\usepackage{graphicx}
\usepackage[utf8]{inputenc} 
\usepackage[T1]{fontenc}    
\usepackage{url}            
\usepackage{booktabs}       
\usepackage{amsfonts}       
\usepackage{nicefrac}       
\usepackage{microtype}      
\usepackage{amssymb}
\usepackage{amsmath}
\usepackage{bbm}

\usepackage{algorithm}
\usepackage{algorithmic}
\usepackage{subfigure}
\usepackage{epstopdf}
\usepackage{booktabs}
\usepackage{array}
\usepackage{multirow}
\usepackage{hhline}
\usepackage{makecell}
\usepackage{framed}
\usepackage{enumitem}
\usepackage{textcomp}

\usepackage{xcolor}

\makeatletter
\newcommand{\printfnsymbol}[1]{%
	\textsuperscript{\@fnsymbol{#1}}%
}
\makeatother

\usepackage[pagebackref=true,breaklinks=true,colorlinks,bookmarks=false]{hyperref}

\def\cvprPaperID{2359} 
\def\confYear{CVPR 2022}

\graphicspath{{images/}}

\begin{document}

\title{Multi-Robot Active Mapping via Neural Bipartite Graph Matching}

\author{
	Kai Ye$^{1}$\thanks{Joint first authors} \quad Siyan Dong$^{2,1}$\printfnsymbol{1} \quad Qingnan Fan$^{3}$\thanks{Corresponding authors} \quad He Wang$^{1}$ \quad Li Yi$^{4}$ \quad Fei Xia$^{5}$ \\ \quad Jue Wang$^{3}$ \quad Baoquan Chen$^{1}$\printfnsymbol{2} \\
	$^1${Peking University} \quad $^2${Shandong University} \quad $^3${Tencent AI Lab} \\ \quad $^4${Tsinghua University} \quad $^5${Stanford University}\\
	{\tt\small \{siyandong.3, fqnchina, ericyi0124, xf1280, arphid, baoquan.chen\}@gmail.com}\\
	{\tt\small \{ye\_kai, hewang\}@pku.edu.cn}
}

\maketitle

\begin{abstract}
We study the problem of multi-robot active mapping, which aims for complete scene map construction in minimum time steps. The key to this problem lies in the goal position estimation to enable more efficient robot movements. Previous approaches either choose the frontier as the goal position via a myopic solution that hinders the time efficiency, or maximize the long-term value via reinforcement learning to directly regress the goal position, but does not guarantee the complete map construction. In this paper, we propose a novel algorithm, namely NeuralCoMapping, which takes advantage of both approaches. We reduce the problem to bipartite graph matching, which establishes the node correspondences between two graphs, denoting robots and frontiers. We introduce a multiplex graph neural network (mGNN) that learns the neural distance to fill the affinity matrix for more effective graph matching. We optimize the mGNN with a differentiable linear assignment layer by maximizing the long-term values that favor time efficiency and map completeness via reinforcement learning. We compare our algorithm with several state-of-the-art multi-robot active mapping approaches and adapted reinforcement-learning baselines. Experimental results demonstrate the superior performance and exceptional generalization ability of our algorithm on various indoor scenes and unseen number of robots, when only trained with 9 indoor scenes.
\end{abstract}

\section{Introduction}

Constructing the map of indoor environments is of great importance to a wide range of applications in the computer vision and robotics communities. With the fast development of range sensors (Kinect, RealSense), many scene mapping approaches \cite{izadi2011kinectfusion,newcombe2011kinectfusion,chen2013scalable,dai2017bundlefusion} are developed to empower scene traversal by human operators with handheld sensors, yet incomplete or unaligned scene meshes are common flaws for inexperienced users due to the noisy and unstable scanned trajectory. To alleviate the inconvenience of human-operated traversal, there emerges autonomous map construction \cite{yamauchi1997frontier,dornhege2013frontier,kim2015active,rodriguez2018importance} via active sensor movement, also known as \textit{active mapping}. 
Previous works in this field mainly focus on using a single robot, which is time-consuming for large-scale environments. In this paper, we study the problem of \textit{multi-robot active mapping}: coordinating multiple robots for the autonomous reconstruction of unknown scenes.

The goal of active mapping is mainly twofold: \textit{time efficiency}, and \textit{map completeness}. The pioneering work for active mapping \cite{yamauchi1997frontier} introduces the concept of frontier: regions on the boundary between open space and unexplored space. 
By continuously moving the robot to new frontiers, the scene map can be completely constructed when no frontier can be found. Many follow-up approaches in the following decades \cite{cao2021tare,dong2019multi,ramakrishnan2020occupancy} aim to improve the time efficiency of the process. However, the problem of active mapping is highly ambiguous, which makes a theoretically-optimal solution almost impossible to be found in an unknown environment.

The key module of active mapping that influences time efficiency is the global planner that estimates the goal position for path planning. The vast majority of literature for both single robot \cite{bourgault2002information,stachniss2005information,tabib2016computationally,bai2016information} and multiple robots \cite{burgard2005coordinated,nieto2014coordination,corah2017efficient,dong2019multi} are \textit{frontier-based}, which decides the goal position from a set of frontiers. However, these approaches are mostly myopic \cite{cao2021tare} and hence hinder the time efficiency, since they either handcraft heuristics \cite{yamauchi1997frontier,dornhege2013frontier,holz2010evaluating} to choose the frontier in the shortest geodesic distance to the robot, or find the one that maximizes the information gain over the next few actions via information-theoretic optimization \cite{stachniss2005information,bai2016information}.
The more recent approaches adopt the \textit{reinforcement learning} strategies \cite{chen2019learning,chaplot2020learning,ramakrishnan2020occupancy} as a replacement of the traditional approaches to decide the goal position for single robot. These policy learning approaches have dominated the active mapping field lately, thanks to their potential to achieve more efficient solutions by maximizing the long-term value \cite{huang2019coloring,kool2018attention}. However, as their goal positions are mostly regressed and may not lay on the frontiers, it has no guarantee to construct the complete map \cite{chen2019learning,chaplot2020learning}.
When the setting of active mapping is extended to the multi-robot scenario, the action space is linearly increased with the robot number, which makes the problem even more ambiguous. The past multi-robot approaches \cite{burgard2005coordinated,nieto2014coordination,corah2017efficient,dong2019multi,faigl2013determination} are mostly frontier-based myopic solutions and are still limited in time efficiency.

In this paper, we propose a novel multi-robot active mapping approach that takes advantage of both the traditional frontier-based and recent reinforcement learning solutions for more efficient and complete map construction. To be specific, we coordinate multiple robots to decide the goal positions from a set of frontiers according to the neural distance optimized by maximizing the long-term value via reinforcement learning. To achieve this goal, we reduce the multi-robot active mapping problem to \textit{bipartite graph matching}, which establishes node correspondences between two graphs, denoting robots and frontiers separately. The key issue for bipartite graph matching lies in the computation of the affinity matrix between two sets of nodes. The traditional frontier-based approaches can be considered as handcrafting the affinity matrix with the geodesic distance between robots and frontiers, which limits the time efficiency of active mapping. In our algorithm, we propose to learn the neural distance with a \textit{multiplex graph neural network} (mGNN) to estimate the affinity matrix for graph matching. 
The problem of graph matching is NP-hard in nature \cite{caetano2009learning} and often formulated as quadratic assignment programming, which is expensive and complex to solve. Many recent works relax graph matching as a linear assignment problem \cite{wang2019learning}, which can be efficiently tackled with a differentiable and approximate solution \cite{sinkhorn1967concerning}. Therefore, we optimize the graph neural network with the differentiable linear assignment by maximizing the long-term value that favors high time efficiency and map completeness via reinforcement learning.

Our algorithm is trained with only 9 indoor scenes, and exhibits exceptional generalization ability to various indoor scene datasets and unseen number of robots. The experimental results demonstrate the superiority of our algorithm over state-of-the-art multi-robot active mapping approaches and a couple of adapted reinforcement-learning baselines. 

All in all, our contributions can be summarized as follows:
\begin{itemize}[nosep]
\item We reduce the multi-robot active mapping problem to bipartite graph matching, which is solved by a novel multi-robot active mapping algorithm that takes advantage of both the traditional frontier-based and recent reinforcement learning approaches.

\item Our algorithm employs a multiplex graph neural network to estimate the affinity matrix, followed by a linear assignment layer for graph matching. The entire process is optimized by maximizing the long-term value via reinforcement learning.

\item While achieving the complete map construction, our algorithm outperforms the existing multi-robot active mapping approaches over time efficiency by a large margin, and demonstrates exceptional generalization ability to unseen robot numbers.

\end{itemize}

\begin{figure*}[t]
\centering
	\includegraphics[width=0.99\linewidth]{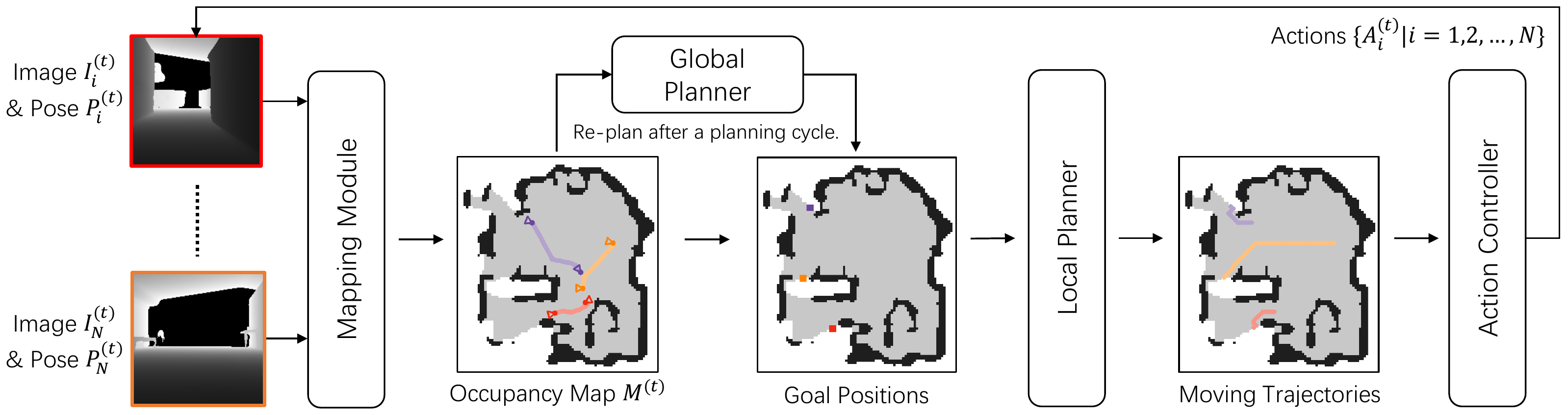}
	\caption{
	    \small{The entire pipeline of our framework. In each planning cycle, mapping (Section \ref{sub_sec:mapping}), local planning and action control (Section \ref{sub_sec:local_action}) are conducted at each step, while global planning (Section \ref{sub_sec:global}) is conducted only once at the beginning. Such a planning cycle is iterated until meeting the termination criterion of active mapping.}
	}
	\label{figure:framework}
	\vspace{-5mm}
\end{figure*}

\section{Related Works}

\textbf{Single-robot active mapping.} 
The vast majority of past works for active mapping lie in the single-robot scenario \cite{yamauchi1997frontier,dornhege2013frontier,holz2010evaluating,kulich2011distance}. The pioneering work of Yamauchi \cite{yamauchi1997frontier} for active mapping presents the concept of frontier and moves the robot towards the nearest frontier in the occupancy map. Dornhege and Kleiner \cite{dornhege2013frontier} extend this idea with the new concept of void, which is the unexplored volumes in the 3D occupancy map. 
The other popular thread for active mapping relies on the information theory \cite{bourgault2002information,stachniss2005information,tabib2016computationally,bai2016information} to choose the frontier based on instant information gain. 

Unlike the above traditional approaches that decide the goal position from a set of frontiers mainly with myopic strategies, recent approaches \cite{chen2019learning,chaplot2020learning} directly employ a convolution neural network to regress the goal position by maximizing the long-term value via reinforcement learning. 
Ramakrishnan et al. \cite{ramakrishnan2020occupancy} follow the above works and proposes to broaden the map coverage beyond the visible area through occupancy anticipation. Some other works explore the environment by constructing a topological map \cite{chaplot2020neural} or semantic map \cite{chaplot2020object}, which is beyond the scope of this work. 

\textbf{Multi-robot active mapping.} 
The large variety of related works about multi-robot active mapping \cite{dong2019multi,visser2013discussion,bhattacharya2014multi,faigl2012goal,burgard2005coordinated,nieto2014coordination,corah2017efficient,yu2021learning} share a lot of methodologies with the single-robot approaches and differ mainly in how they coordinate multiple robots for the goal assignment. 
Faigl et al. \cite{faigl2012goal} have a good summary of the common multi-robot active mapping approaches.
One solution \cite{werger2000broadcast} is to sort the robot-goal pairs with the geodesic distance and traverse the ordered sequence from the first element to assign the next not-assigned goal to the robot. 
More recent approaches \cite{dong2019multi,faigl2012goal} mainly rely on solving an optimal mass transport problem between robots and goals, such as multiple traveling salesman problem, for goal assignments.
Most of the aforementioned approaches are myopic as analyzed \cite{dong2019multi,cao2021tare} since the goal positions are chosen from the nearest target in the geodesic distance with multi-robot coordination constraints. In this work, we propose to learn better neural distance via reinforcement learning to achieve more efficient map construction. 

\textbf{Graph neural networks.}
Graph neural networks \cite{wu2020comprehensive} enable learning on top of graph representations. Sykora et al. \cite{sykora2020multi} use graph neural network to solve the multi-agent graph coverage problem. Zhang et al. \cite{zhang2020multiplex} introduce multiplex network structures for the multi-behavior recommendation. In this work, we perform learning on multiple graphs, namely multiplex graph neural network (mGNN), to estimate the affinity matrix for bipartite graph matching.

\section{Problem Statement}

In an unknown environment, the goal of multi-robot active mapping is to construct a complete map in minimum time steps. At each time step $t$, the robot $i$ receives a first-person depth image $I^{(t)}_i$ and its corresponding camera pose $P^{(t)}_i$ as input, and estimates its action $A^{(t)}_i$ for movement, following the problem setting of the previous work closest to ours \cite{dong2019multi}. We adopt the common TurtleBot model as our robot, and it runs in the physically-realistic simulator iGibson \cite{shen2021igibson,li2021igibson}, which contains the physical robot body and simulates the realistic action noise and collisions.

For better comparison with previous works \cite{dong2019multi,faigl2012goal}, we further consider the problem settings below. 
1) \textit{Map completeness first}: we value map completeness the most, hence the map is continuously explored until no accessible frontier is found.
2) \textit{Co-located robots}: the pose information is shared among all the robots, hence the global map can be constructed by synchronizing the local maps from all the robots. 3) \textit{Spatially-close initialization}: all the robots are randomly initialized in the traversable region of the map with the constraint that the geodesic distance between every two robots is smaller than a threshold $\lambda_r$. Note uniformly sampling the robots in the entire map will save more scanning effort for exploration, and we are working in a more challenging setting with spatially-close robot initialization.


\begin{figure*}[t]
\centering
	\includegraphics[width=0.99\linewidth]{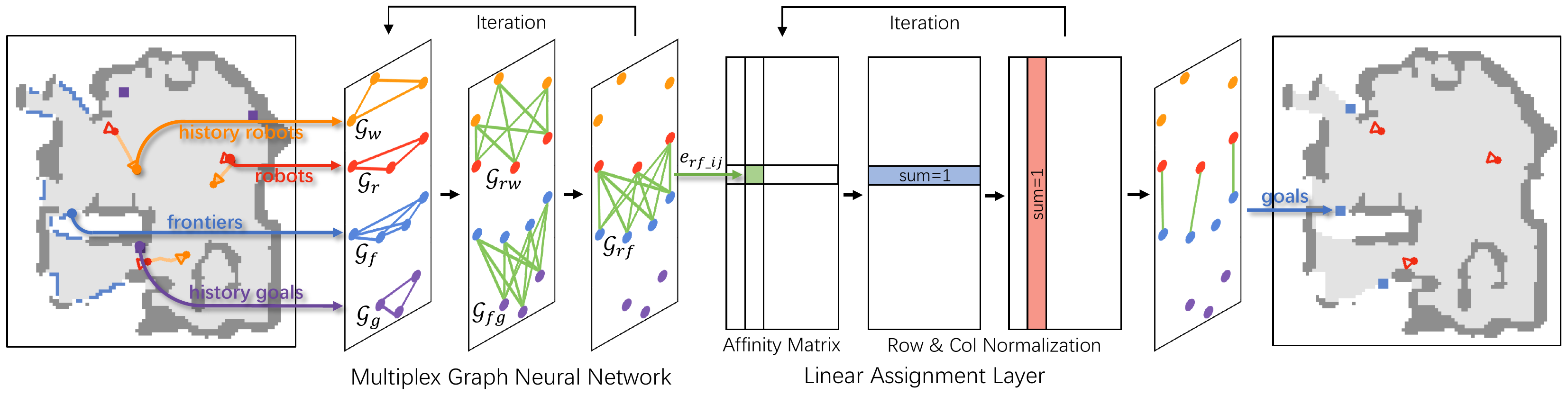}
	\caption{
	    \small{Illustration of the global planner. It consists of a multiplex graph neural network (mGNN) that learns the affinity information between robots and frontiers, and a linear assignment layer to match them. The robot and frontier nodes are extracted from the mapping module, while the history robot and goal nodes are extracted from the past planning cycles. With the affinity matrix formed by the learned directed edge features, the linear assignment layer conducts bipartite graph matching to assign each robot a unique frontier as the goal.}
	}
	\label{figure:global}
	\vspace{-3mm}
\end{figure*}

\section{NeuralCoMapping}
We introduce the entire framework (Figure \ref{figure:framework}) of our algorithm below. 
The \textit{mapping module} constructs the occupancy map based on the current depth observations and camera poses from all the robots. The \textit{global planner} estimates the goal position for each robot in the occupancy map by solving a neural bipartite graph matching problem. To navigate to the goal position, the \textit{local planner} calculates an obstacle-free moving trajectory for each robot, followed by the \textit{action controller} that performs the specific robot moves along the trajectory. 
We define a \textit{planning cycle} as a short period of a fixed horizon. The global planner only estimates the goal positions at the beginning of the planning cycle, while mapping, local planning, and action control are alternatively conducted until the end of the planning cycle. Such a planning cycle is iterated until meeting the termination criterion.

\subsection{Mapping Module} \label{sub_sec:mapping}
Provided the depth observations and the corresponding camera poses from all the robots, the goal of the mapping module is to construct the global map.
We represent the map at time step $t$ as the occupancy grid, which is the top-down view of the 3D scene, denoted as $M^{(t)} \in [0,1]^{X \times Y \times 2}$, where $X,Y$ are the map size. The two map channels indicate the explored and occupied regions separately. Hence each cell in the grid can be classified as one of the three classes, open (explored but not occupied), occupied, and unknown (unexplored). Frontier is defined as the open cells whose adjacency contains unknown cells. The occupancy map $M^{(0)}$ is unexplored and initialized as all zero. For robot $i$, the pose of its mounted camera $P^{(t)}_i \in \mathbb{SE}(2)$ is represented as $(x,y,\theta)$, where $x\in [0, X-1],y\in [0, Y-1]$ denote the robot position and $\theta \in [0, 2\pi)$ denotes the robot orientation around the vertical axis.

To compute the occupancy map, the depth image $I^{(t)}_i$ is firstly back-projected as the point cloud in the world space with known camera intrinsics and provided poses $P^{(t)}_i$, then the 3D points from all the robots are fused and projected onto the 2D plane to overlay the overlapping region of the existing occupancy map $M^{(t-1)}$ to obtain the new one $M^{(t)}$.
We consider a cell as occupied if there is any 3D point (staying between two horizontal planes at the height of the ground and robot head) that falls into the cell, and consider a cell as open if it lies on the ray from the camera center to the end of the definitively visible space \cite{hane2015obstacle}.
The multi-robot scenario creates a highly dynamic environment, where one moving robot may be visually observed by the others. It poses new challenges for obstacle-free path planning. To tackle this issue, we label all the cells covered by any robot being observed as occupied, and label those cells as open again once the robot leaves them.

\subsection{Global Planner} \label{sub_sec:global}
With the constructed occupancy map and provided robot positions, the global planner aims to estimate the optimal goal positions for all the robots for efficient and complete map construction. 
In this work, we formulate the problem of goal position estimation as \textit{bipartite graph matching}, which establishes the correspondences between the robot and frontier\footnote{The frontier nodes are sampled from the frontier cells, and each frontier node represents a frontier cell.} nodes extracted from the constructed occupancy map. 
The affinity matrix for graph matching is not composed of the geodesic distance as adopted in the traditional frontier-based approaches \cite{visser2013discussion,bhattacharya2014multi,faigl2012goal,dong2019multi}, yet is filled with the neural distance estimated by a graph neural network, which is optimized with a differentiable linear assignment layer by maximizing the long-term value via reinforcement learning. 
We introduce the solution to bipartite graph matching via the following two components, \textit{multiplex graph neural network} for affinity matrix estimation, and \textit{linear assignment layer} to pair the robot and frontier nodes. Figure \ref{figure:global} gives the illustration of the aforementioned global planner.

\subsubsection{Multiplex Graph Neural Network}

Provided the constructed occupancy map and known robot positions, we first construct two self-graphs $\mathcal{G}_r = (\mathcal{V}_r, \mathcal{E}_r)$ and $\mathcal{G}_f = (\mathcal{V}_f, \mathcal{E}_f)$ that denote the robot and frontier sets separately. We also build a cross-graph $\mathcal{G}_{rf} = (\mathcal{V}_r, \mathcal{V}_f, \mathcal{E}_{rf})$ that connects the robots and frontiers, and $\mathcal{E}_{rf}$ denotes the affinity information we want to learn for graph matching.
The multiplex graph neural network (mGNN) learns such information between robots and frontiers with iterative intra-graph and inter-graph operations.

For sake of simplicity, we introduce the node feature computation only for robots below, and it applies to the frontier nodes as well. For robot $i$, we represent the raw robot information as $s_{r\_i} \in \mathbb{R}^3$, which includes the $x,y$ coordinates in the occupancy map and its semantic label (\textit{robot}, or \textit{frontier}). We extract the initial high-dimensional robot node feature $v^{(0)}_{r\_i} \in \mathbb{R}^{32}$ from $s_{r\_i}$ via a multi-layer perception (MLP) $f_{init}$:
\begin{equation}
v^{(0)}_{r\_i} = f_{init}(s_{r\_i})
\end{equation}

\textbf{Intra-graph operation.}
The intra-graph operation updates the node and edge features for both the robot and frontier self-graphs.
We consider a fully connected self-graph, hence each node will be updated by the messages received from all the other nodes. Inspired by the attention mechanism \cite{vaswani2017attention}, such a node aggregation operation can be treated as the retrieval process which maps your query against a set of keys associated with candidate nodes in the graph and finally presents the best matched nodes (values). Hence, we first compute the query $q^{(l)}_{r\_i}\in \mathbb{R}^{32}$, key $k^{(l)}_{r\_i}\in \mathbb{R}^{32}$, and value $u^{(l)}_{r\_i}\in \mathbb{R}^{32}$ from the node feature $v^{(l)}_{r\_i}$ at layer $l$,
\begin{equation}
q^{(l)}_{r\_i} = f_{query}(v^{(l)}_{r\_i}),
k^{(l)}_{r\_i} = f_{key}(v^{(l)}_{r\_i}),
u^{(l)}_{r\_i} = f_{value}(v^{(l)}_{r\_i})
\end{equation}
where $f_{query},f_{key},f_{value}$ are parameterized as linear projections. Then we represent the directed edge feature $e^{(l)}_{r\_{ij}} \in \mathbb{R}^{1}$ from node $j$ to node $i$ as the attention weight scalar,
\begin{equation}
e^{(l)}_{r\_{ij}} = \frac{\exp{(q^{(l)}_{r\_i} \cdot k^{(l)}_{r\_j})}}{\sum_{h:(i,h) \in \mathcal{E}_r} \exp{(q^{(l)}_{r\_i} \cdot k^{(l)}_{r\_h})}}
\end{equation}
which is the softmax over all the query-key dot product results directed to node $i$. Then the node feature $v^{(l+1)}_{r\_i}$ is computed as
\begin{equation} \label{equation:self_node}
v^{(l+1)}_{r\_i} = v^{(l)}_{r\_i} + f_{node}(v^{(l)}_{r\_i}, \sum_{h:(i,h) \in \mathcal{E}_r} e^{(l)}_{r\_{ih}} u^{(l)}_{r\_h})
\end{equation}
where $f_{node}$ concatenates the input information and is parameterized as a multi-layer perception.

\textbf{Inter-graph operation.}
The inter-graph operation updates the node and edge features for the cross-graph. 
We consider a complete bipartite graph, where each node in one set is connected with all the nodes in the other set, yet not connected with any nodes in the same set.
The geodesic distance is the shortest distance for traversal between two points and implicitly encodes the underlying scene layout information. We compute geodesic distances with the Fast Marching Method (FMM) \cite{sethian1996fast}. 
The geodesic distance between two nodes $i,j$ is denoted as $d_{ij}$, which is incorporated into the directed edge feature $e^{(l)}_{rf\_{ij}} \in \mathbb{R}^{1}$ computation as below,
\begin{equation}
e^{(l)}_{rf\_{ij}} = \frac{\exp{(f_{edge}(q^{(l)}_{r\_i},k^{(l)}_{f\_j},d_{ij}))}}{\sum_{h:(i,h) \in \mathcal{E}_{rf}} \exp{(f_{edge}(q^{(l)}_{r\_i},k^{(l)}_{f\_h},d_{ih}))}}
\end{equation}
where $f_{edge}$ concatenates the input information and is parameterized as a multi-layer perception. Then the node feature $v^{(l+2)}_{r\_i}$ is computed similarly to Equation \ref{equation:self_node} by replacing the directed edges in $\mathcal{E}_r$ with $\mathcal{E}_{rf}$,
\begin{equation}\label{equation:node_aggreg_inter}
v^{(l+1)}_{r\_i} = v^{(l)}_{r\_i} + f_{node}(v^{(l)}_{r\_i}, \sum_{h:(i,h) \in \mathcal{E}_{rf}} e^{(l)}_{rf\_{ih}} u^{(l)}_{f\_h})
\end{equation}

\textbf{History node module.}
For the problem of multi-robot active mapping, not only do the current robots and frontiers matter for the global planning, the robots and estimated goals in the past should also serve as the guidance to encourage consistent robot movements and discourage the redundant traversal over explored regions.
Therefore, we further enhance mGNN with two sets of new nodes with the semantic label of \textit{history robot} or \textit{history goal}, which are derived from the robots and goals in the past individually. 
To be specific, we construct two more self-graphs $\mathcal{G}_w = (\mathcal{V}_w, \mathcal{E}_w)$, $\mathcal{G}_g = (\mathcal{V}_g, \mathcal{E}_g)$ that denote the history robots and history goals, and two more cross-graphs $\mathcal{G}_{rw} = (\mathcal{V}_r,\mathcal{V}_w, \mathcal{E}_{rw})$, $\mathcal{G}_{fg} = (\mathcal{V}_f,\mathcal{V}_g, \mathcal{E}_{fg})$ that associate robots with history robots, frontiers with history goals separately. In this manner, we achieve the history node module by applying the intra-graph operation to $\mathcal{G}_g,\mathcal{G}_w$, and the cross-graph operation to $\mathcal{G}_{fg},\mathcal{G}_{rw}$ as well. 
In the implementation, instead of considering the history robot positions among all the past steps, we only count in the history robot positions at the beginning of each past planning cycle to balance the number of history robot and history frontier nodes.

Therefore, the entire mGNN is composed of four self-graphs $\mathcal{G}_r, \mathcal{G}_f, \mathcal{G}_w, \mathcal{G}_g$ and three cross-graphs $\mathcal{E}_{rf}, \mathcal{E}_{rw}, \mathcal{E}_{fg}$. For each intra-graph operation, all the four self-graphs can be simultaneously updated, while for each inter-graph operation, the three cross-graphs need to be sequentially updated as the robot and frontier nodes will be updated twice in two cross-graphs. Note the cross-graph order for inter-graph operation does not affect the final node feature, which is simply added with the residual as computed in Equation \ref{equation:node_aggreg_inter}.
We consider one block of graph operations as the composition of one intra-graph and one inter-graph operation. The entire mGNN is composed of $N_l$ blocks of graph operations.

\subsubsection{Linear Assignment Layer} 

For our bipartite graph matching problem, the affinity matrix denotes the distance between the robot and frontier nodes. The learned edge feature $e_{rf\_{ij}}$ emitted from robot $j$ to frontier $i$ indicates how much the robot prefers the frontier and is treated as the neural distance to form the affinity matrix for graph matching\footnote{Experimentally we observe the directed edge feature from the frontier to robot makes the similar effect.}. With the cross-graph $\mathcal{G}_{rf} = (\mathcal{V}_r, \mathcal{V}_f, \mathcal{E}_{rf})$, the goal of multi-robot active mapping can be considered as achieving a maximum matching in a bipartite graph. It is required to conduct any many connections between robot and frontier nodes as possible by assigning at most one robot to one frontier, and also at most one frontier to one robot, in such a way the summed affinity among all the connections are maximized. It corresponds to the linear assignment problem and can be solved efficiently with the popular Sinkhorn algorithm \cite{sinkhorn1967concerning}. It works by normalizing the rows and columns of the affinity matrix alternatively until convergence and is often treated as the approximate and differentiable version of the Hungarian algorithm \cite{munkres1957algorithms}. We implement the linear assignment layer as the Sinkhorn algorithm, whose output decides the goal position as the matched frontier for each robot.

\subsection{Local Planner and Action Controller} \label{sub_sec:local_action}
With the robot position, estimated goal position, and constructed occupancy map, the local planner decides a moving trajectory from the robot to the goal position. We adopt Fast Marching Method (FMM) \cite{sethian1996fast} to achieve this purpose. However, due to the fact that the occupancy map is only a rough representation of the scene world and the unavoidable action noise when executing the moving trajectory, the robot will collide with obstacles from time to time, for example, in the narrow corridor, in which situation the fast marching approach may not help free the robot from the collision in the physically-realistic simulator iGibson \cite{shen2021igibson,li2021igibson} and hence time efficiency is significantly influenced. To alleviate the above difficulty, we propose to improve FMM with an \textit{obstacle-resistant strategy} that plans the moving trajectory away from the obstacles. 
To be specific, given the occupancy map and robot positions at time step $t$, we first generate two geodesic distance maps $D^{(t)}_{r}, D^{(t)}_{o}$ whose zero contour lies in the robot positions and obstacles separately with FMM, then update the robot distance value $D^{(t)}_{r\_i}$ at position $i$ as,
\begin{equation}
D^{(t)}_{r\_i} = D^{(t)}_{r\_i} / (\epsilon + min(\tau, D^{(t)}_{o\_i}) \times \lambda_o)
\end{equation}
where $\tau = 0.001, \tau = 4, \lambda_o = 0.25$. It means amplifying the robot distance values whose positions are close to obstacles. Then $D^{(t)}_{r}$ is used to calculate the moving trajectory from the robot to its goal position.

The robot is able to perform three actions, $A = \{{move\_forward}, {turn\_left}, {turn\_right}\}$. Given the next waypoint in the moving trajectory, the robot controls its actions via a simple heuristic \cite{chen2021topological}: if the robot faces the waypoint, it moves forward; otherwise, it rotates towards the waypoint. To be specific, 
we compute the relative angle $\theta_a$ by subtracting the robot orientation from the orientation of the directed robot-to-waypoint edge, the action $A^{(t)}_i$ at time step $t$ for robot $i$ is decided as,
\begin{equation}
 A^{(t)}_i=\begin{cases}
      move\_forward & \text{if $\theta_a > -\lambda_a$ or $\theta_a < \lambda_a$} \\
      turn\_left & \text{if $\theta_a \leq -\lambda_a$} \\
      turn\_right & \text{if $\theta_a \geq \lambda_a$}
    \end{cases}
\end{equation}
where $\lambda_a$ is the threshold that controls the forward and rotation movement.

\subsection{Reinforcement Learning Design}
We treat the mGNN as the policy network, which is optimized with the differentiable linear assignment layer together by maximizing the accumulated reward in the entire episode via reinforcement learning. Despite the multi-robot scenario, as our global planner adopts the centralized decision setting, we directly use the off-policy learning approach Proximal Policy Optimization (PPO) \cite{schulman2017proximal} as the policy optimizer.

\textbf{Reward function.}
The goal of active mapping is to pursue high time efficiency and map completeness. To achieve this goal, we design a time reward $\hat{\mathcal{R}}_{time}$ and a coverage reward $\hat{\mathcal{R}}_{coverage}$. The time reward punishes unnecessary time steps to encourage high exploration efficiency, hence is defined as,
\begin{equation}
 \hat{\mathcal{R}}_{time} = -0.015
\end{equation}
Coverage $C(M^{(t)})$ at time step $t$ is the area of the open space in the occupancy map $M^{(t)}$. The coverage reward $\hat{\mathcal{R}}_{coverage}$ is defined as the coverage increment measured in $m^2$,
\begin{equation}
 \hat{\mathcal{R}}_{coverage} = C(M^{(t)})- C(M^{(t-1)})
\end{equation}
Then the entire reward $\hat{\mathcal{R}}$ is computed as the weighted summation of the two above two rewards, 
\begin{equation}
 \hat{\mathcal{R}} = \hat{\mathcal{R}}_{time} + \lambda_c \hat{\mathcal{R}}_{coverage}
\end{equation}
where $\lambda_c$ is the hyper-parameter that balances the two rewards.

\setlength{\tabcolsep}{10pt}
\begin{table*}[t]
\centering
    \small
    \begin{tabular}{l|cc|cc|cc}
        \toprule
        & \multicolumn{2}{c}{Small Scenes ($<35 m^2$)} & \multicolumn{2}{|c}{Middle Scenes ($35-70 m^2$)} & \multicolumn{2}{|c}{Large Scenes ($>70 m^2$)} \\
        \midrule
        {Method} & 
        {Cov. (\%)} &
        {Time (\#steps)} &
        {Cov. (\%)} &
        {Time (\#steps)} &
        {Cov. (\%)} &
        {Time (\#steps)} \\
        \midrule
        Greedy \cite{visser2013discussion} & 98.7 & 420.6 & 98.8 & 669.4 & 99.4 & 1057.0 \\
        VorSEG \cite{bhattacharya2014multi} & 98.5 & 350.8 & 98.8 & 570.3 & 99.6 & 1080.5 \\
        mTSP \cite{faigl2012goal} & 98.8 & 351.4 & 98.6 & 564.5 & 98.9 & 1029.0 \\
        CoScan \cite{dong2019multi} & 98.6 & \textcolor{blue}{304.2} & 99.0 & \textcolor{blue}{496.1} & 99.5 & \textcolor{blue}{985.0} \\
        \midrule
        NeuralCoMapping (Ours) & 98.6 & \textcolor{red}{302.5} (\textcolor{orange}{-0.6\%}) & 98.8 & \textcolor{red}{471.7} (\textcolor{orange}{-4.9\%}) & 98.9 & \textcolor{red}{882.0} (\textcolor{orange}{-10.5\%}) \\
        \bottomrule
    \end{tabular}
    \vspace{2mm}
    \caption{Numerical results on the Gibson dataset \cite{xia2018gibson}. 
    Parentheses: \%steps reduced against the best competitor (blue) for our algorithm (red).
    }
\label{table:Gibson}
\vspace{-3mm}
\end{table*}

\setlength{\tabcolsep}{8pt}
\begin{table*}[t]
\centering
    \small
    \begin{tabular}{l|cc|cc|cc}
        \toprule
        & \multicolumn{2}{c}{Small Scenes ($<100 m^2$)} & \multicolumn{2}{|c}{Middle Scenes ($100 - 300 m^2$)} & \multicolumn{2}{|c}{Large Scenes ($>300 m^2$)} \\
        \midrule
        {Method} & 
        {Cov. (\%)} &
        {Time (\#steps)} &
        {Cov. (\%)} &
        {Time (\#steps)} &
        {Cov. (\%)} &
        {Time (\#steps)} \\
        \midrule
        Greedy \cite{visser2013discussion} & 95.9 & 801.2 & 95.1 & 2043.1 & 91.3 & 3345.4 \\
        VorSEG \cite{bhattacharya2014multi} & 95.5 & 652.7 & 94.7 & 1693.2 & 91.0 & 2852.0 \\
        mTSP \cite{faigl2012goal} & 96.6 & 712.5 & 95.6 & 1742.6 & 91.4 & 2963.6 \\
        CoScan \cite{dong2019multi} & 97.1 & \textcolor{blue}{581.7} & 96.1 & \textcolor{blue}{1505.5} & 92.1 & \textcolor{blue}{2781.0} \\
        \midrule
        ANS-DeCen & 85.1 & 1860.4 & 59.9 & 3229.8 & 48.5 & 5639.6 \\
        ANS-Cen & 89.7 & 1536.8 & 64.3 & 2781.3 & 52.2 & 4961.3 \\
        \midrule
        NeuralCoMapping (Ours) & 96.7 & \textcolor{red}{506.1} (\textcolor{orange}{-13.0\%}) & 96.0 & \textcolor{red}{1217.3} (\textcolor{orange}{-19.1\%}) & 92.4 & \textcolor{red}{1874.8} (\textcolor{orange}{-32.6\%}) \\
        \bottomrule
    \end{tabular}
    \vspace{2mm}
    \caption{Generalization to the unseen Matterport3D dataset \cite{chang2017matterport3d}, which is consistently larger than the Gibson dataset. 
    Note our algorithm is trained only on 9 scenes in the Gibson dataset, while the ANS variants are trained on the entire Gibson dataset.
    }
\label{table:Matterport3D}
\vspace{-3mm}
\end{table*}

\section{Experiments}

\subsection{Experimental Setup}

\textbf{Data processing.} 
Our algorithm is trained on the Gibson dataset \cite{xia2018gibson} and evaluated on both the Gibson and Matterport3D datasets \cite{chang2017matterport3d}. Our algorithm runs in the physically-realistic iGibson \cite{shen2021igibson,li2021igibson} simulator. We adopt the TurtleBot model as our robot, which has a physical body and can be visually observed in the simulator to create a more realistic multi-robot scenario. 
To demonstrate the robustness of our algorithm on training scenes and the extraordinary generalization ability of our algorithm to novel scenes, we collect only 9 scenes randomly sampled in the Gibson dataset for training\footnote{The training is once for all and takes about 1 day.} and use the other scenes for evaluation. 

\textbf{Termination criterion.} 
We aim for complete map construction, however, in the iGibson environment, all the robots have physical bodies and get stuck occasionally \cite{ye2021auxiliary}.
Hence following \cite{dong2019multi}, our algorithm terminates when there is no accessible frontier in the environment.

\textbf{Parameter setting.} 
$X= 480$; $Y = 480$; $N_l = 3$; $\lambda_a = 12.5$ degrees; 
$\lambda_r = 3$ meters; $\lambda_c = 0.005$. Each cell in the grid represents a $0.01m^2$ region. The bandwidth between robots is 10-20 KB. The ${move\_forward}$ action advances 6.5 cm, and the $turn\_left/turn\_right$ action rotates 12.5 degrees. The horizon of the planning cycle is 25 steps. 

\textbf{Evaluation metrics.}
We evaluate the map completeness via the coverage ratio (Cov. (\%)) that calculates the percentage of explored open space over the ground truth open space in the environment. We measure the time efficiency (Time (\#steps)) as the number of steps taken for map construction.

\subsection{Compared Approaches} 

We compare our algorithm with several state-of-the-art multi-robot active mapping approaches (Greedy, VorSEG, mTSP, CoScan) \cite{dong2019multi,visser2013discussion,bhattacharya2014multi,faigl2012goal}, which are mostly frontier-based and rely on the geodesic distance to choose the goal position. Inspired by the past reinforcement learning solutions for single-robot active mapping \cite{chen2019learning,chaplot2020learning,ramakrishnan2020occupancy}, we also design two multi-robot baselines (ANS-DeCen, ANS-Cen) that directly regress the coordinates of goal positions. To justify the effectiveness of our global planner, we compare with these approaches by only replacing our global planner with the alternatives in these approaches while leaving the mapping, local planner, action controller and termination criterion the same for everyone. We elaborate on all these methods below.

\begin{itemize}[leftmargin=0.4cm,nosep]
\item \textbf{Greedy} \cite{visser2013discussion}. It assigns the frontiers to all the robots in a greedy manner. Each robot chooses the closest frontier among its assigned ones as the goal position.

\item \textbf{VorSEG} \cite{bhattacharya2014multi}. It performs a Voronoi segmentation over all the frontiers. Each robot chooses the closest segmented region and scans its frontiers in a greedy manner.

\item \textbf{mTSP} \cite{faigl2012goal}. It constructs a frontier graph where the edge weights are defined as the geodesic distance between two frontiers, then it solves a multiple traveling salesman problem based on the frontier graph for the goal assignment.

\item \textbf{CoScan} \cite{dong2019multi}. It performs a k-means clustering over all the frontiers and assigns each robot a frontier cluster by solving an optimal mass transport problem based on the geodesic distance. Then for each robot, it computes an optimal traverse path over its assigned frontiers.

\item \textbf{ANS-DeCen}. Following the recent reinforcement learning solution, Active Neural SLAM (ANS) \cite{chaplot2020learning}, we replace our global planner with their global policy, which learns a convolution neural network that consumes an egocentric local map and a geocentric global map as input, and regresses the goal estimation for path planning. We design a fully decentralized setting where multiple robots construct the map independently without communication.

\item \textbf{ANS-Cen}. We design a simple centralized setting of the ANS \cite{chaplot2020learning} approach. 
To adapt to the global policy, we modify the input by stacking the local maps together with the information of all robots to fuse the global map and reform the output by regressing the goal positions of all the robots together.

\end{itemize}

\begin{figure*}[t]
\centering
 	\includegraphics[width=0.99\linewidth]{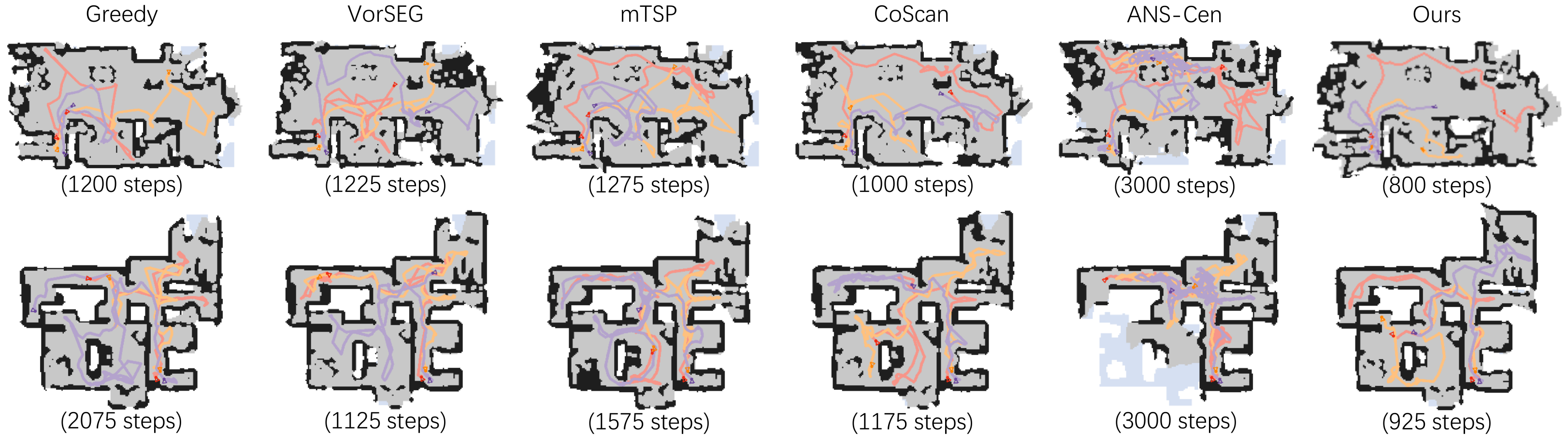}
	\caption{
	    \small{Visual results of our algorithm compared with state-of-the-art approaches on two scene samples from the Matterport3D dataset. 
	    }
	}
	\label{figure:trajectory}
	\vspace{-2mm}
\end{figure*}

\setlength{\tabcolsep}{10pt}
\begin{table*}[t]
\centering
    \small
    \begin{tabular}{c|cc|cc|cc}
        \toprule
        & \multicolumn{2}{c|}{Train with 2 robots} & \multicolumn{2}{c|}{Train with 3 robots} & \multicolumn{2}{c}{Train with 4 robots} \\
        \midrule
        Test number of robots & {Cov. (\%)} & {Time (\#steps)} & {Cov. (\%)} & {Time (\#steps)} & {Cov. (\%)} & {Time (\#steps)} \\
        \midrule
        2 robots & 96.7 & 1002.6 & 96.4 & 1005.0 & 97.1 & 1019.7 \\
        3 robots & 96.7 & 680.0 & 96.3 & 661.7 & 96.1 & 683.7 \\
        4 robots & 97.1 & 617.2 & 96.7 & 626.3 & 96.7 & 614.6 \\
        \bottomrule
    \end{tabular}
    \vspace{2mm}
        \caption{Generalization to unseen numbers of robots evaluated on the Matterport3D dataset. 
        }
\label{table:number_robots}
\vspace{-3mm}
\end{table*}

\setlength{\tabcolsep}{5pt}
\begin{table}[t]
\centering
    \small
    \begin{tabular}{l|cc}
        \toprule
        {Method} & 
        {Cov. (\%)} &
        {Time (\#steps)} \\
        \midrule
        Affinity: geodesic distance & 96.9 & 785.5 \\
        Affinity: node correlation & 94.3 & 1045.6 \\
        w/o history module & 96.1 & 729.5 \\
        w/o obstacle-resistance & 96.8 & 792.7 \\
        \midrule
        NeuralCoMapping (Ours) & 96.3 & 661.7 \\
        \bottomrule
    \end{tabular}
    \vspace{2mm}
    \caption{Ablation study on the Matterport3D dataset.}
\label{table:ablation}
\vspace{-7mm}
\end{table}

\subsection{Results}

We run all the approaches in the 3-robot scenario. The learnable approaches (ANS variants and our approach) are both trained and evaluated with 3 robots. 
The two ANS variants directly extract features from the raw map with a convolution neural network, which is more sensitive to the variance of the map distribution and needs to be trained in a larger set of training scenes (72 scenes in ANS \cite{chaplot2020learning}), unlike 9 scenes in our setting. Hence it would be unfair for the ANS variants to directly compare to our approach with our Gibson train/test split. During the implementation, we choose to train the ANS variants on the entire Gibson dataset and compare with them directly on the Matterport3D dataset.

We demonstrate the results of our algorithm and the four multi-robot active mapping approaches on the Gibson dataset in Table \ref{table:Gibson}. All the approaches are able to achieve the roughly complete map construction, as also observed in \cite{dong2019multi}, while our algorithm achieves superior performance over all the other approaches regarding time efficiency. The scenes in the Gibson dataset are relatively small, where the time efficiency tends to saturate in the multi-robot scenario, hence the performance difference between our approach and the best competitor is less significant. In Table \ref{table:Matterport3D}, when we evaluate the approaches on the Matterport3D dataset, which contains consistently larger scenes than the ones in the Gibson dataset, our algorithm exhibits more significant superiority compared to the best competitor, especially in the large area interval ($>300 m^2$) we save more than 30\% steps.
We count time step rather than running time for the efficiency evaluation, while we take only 0.04s for each global planning, which is much faster than the best competitor CoScan 0.4s. 
The results also demonstrate the outstanding generalization ability of our algorithm to novel indoor scenes. 
Note the ANS variants do not guarantee a complete map construction as their goal positions are regressed and may not lay on frontiers. Therefore, to obtain a reasonable number for comparison, we set a maximum horizon (5000 steps as default, 10000 steps for scenes $>300m^2$), which is sufficiently large for most approaches to complete the map construction.
We visualize the moving trajectories in Figure \ref{figure:trajectory}.

\subsection{Generalization to Unseen Number of Robots}

We evaluate how our algorithm generalizes to the unseen number of robots in Table \ref{table:number_robots}. Our algorithm is trained with 2, 3, and 4 robots separately, and evaluated with different robot numbers on the Matterport3D dataset. From the results, we observe that our algorithm achieves very close time efficiency to its upper bound (trained and evaluated on the same number of robots). We consider such a good generalization ability mainly stems from three aspects. 1) Our entire framework contains only one learnable module, the global planner, while the other modules are non-learnable and can naturally generalize to the unseen number of robots. 2) The multiplex graph neural network in the global planner decomposes the goal estimation into many small and independent tasks, where the robot nodes are only partially involved in the entire network. 3) For the inter- and intra- graph operations in mGNN, the node feature is updated as the weighted sum of its neighborhood features, and hence relatively invariant to the number of its neighborhood (robots).

\subsection{Ablation Study}

We conduct an ablation study to evaluate the importance of each component of our algorithm to the multi-robot active mapping problem, as shown in Table \ref{table:ablation}. We firstly validate the design of the affinity matrix, by replacing the neural distance (edge feature) with 1) the geodesic distance between robots and frontiers and 2) the node correlation computed as the dot product between robot and frontier node features. We further justify the effectiveness of the history node module in mGNN and the obstacle-resistance strategy in the local planner by removing them separately from the entire framework for ablation study. Experimentally, we observe that our full framework achieves the best time efficiency. One of the major arguments in this paper is that the pure geodesic distance is not the optimal measurement to choose a reasonable goal position, which motivates our algorithm to learn the neural distance via reinforcement learning. Such an argument is validated in the experiment, where our neural distance is superior to the pure geodesic distance by a large margin.

\section{Conclusion}
In this work, we propose a novel multi-robot active mapping algorithm to achieve efficient and complete map construction. We formulate the problem as neural bipartite graph matching, which is solved via the proposed multiplex graph neural network and a differentiable linear assignment layer. The entire framework is optimized by maximizing the long-term value via reinforcement learning.

\textbf{Acknowledgements.} We thank the anonymous reviewers for their valuable comments. This work was supported by NSFC (62161146002).

\section{Appendix}
The appendix provides the additional supplemental material that cannot be included in the main paper due to its page limit:

\begin{itemize}
\item Implementation details.
\item Training with a single scene. 
\item Generalization to more unseen robots. 
\end{itemize}

\section*{A. Implementation details}

We train the global planner with 12 parallel threads. We use 12 mini-batches and do 4 epochs in each PPO update. We adopt the Adam optimizer with a learning rate of 0.0005/0.000025 (actor/critic), a discount factor of 0.99, an entropy coefficient of 0.0001 and value loss coefficient of 3.0.
The network structure for the multiplex graph neural network is detailed below:
\begin{itemize}[nosep]
\item $f_{init}$: a 5-layer MLP (3-32-64-128-256-32).
\item $f_{query}/f_{key}/f_{value}$: a linear projection (32-32).
\item $f_{node}$: a 2-layer MLP (64-64-32).
\item $f_{edge}$: a 2-layer MLP (96-32-1).
\end{itemize}

Each fully connected layer in the above networks is followed by a Batch Normalization layer and a ReLU layer.

In the iGibson environment, each robot has a physical body and can be visually observed by other robots and hence becomes obstacles (occupied) in the occupancy map. Therefore, unlike the previous works \cite{chen2019learning,chaplot2020learning,ramakrishnan2020occupancy} that do not update the obstacle once it is constructed, in our framework, we update the explored region (free and explored region) when it has been scanned more than once to alleviate the issues raised by the multi-robot scenario.

\section*{B. Training with a single scene}

In the main paper, we demonstrate that our algorithm is trained only on 9 scenes in the Gibson dataset, and is able to generalize well to various indoor scenes even in the unseen Matterport3D dataset. This mainly benefits from the proposed multiplex graph neural network, which is the solely learnable module in our algorithm, and only relies on the simple robot, frontier, and their geodesic distance information extracted from the occupancy map for goal position estimation. Such a design makes our algorithm relatively robust to the geometry, appearance, and layout variations of the indoor scene distributions.

To further explore the potential of training with fewer scenes for our algorithm, we experiment with an extreme case, where our algorithm is trained with only a single scene. The results are shown in Table \ref{table:single_scene}. Surprisingly, we observe that in this extreme case, our algorithm still performs comparably with the original one trained with nine scenes. It demonstrates the strong learning ability and robustness of our algorithm.

\section*{C. Generalization to more unseen robots}

We further evaluate the generalization ability of our algorithm on more unseen robots. We train our algorithm with 3 robots, same as the multi-robot scenario in the main paper, and evaluate it with 2, 4, 5, 7 and 9 robots. We also test its upper bound performance by training and evaluating on the same number (2, 4, 5, 7, 9) of robots. We observe that when we run more robots for scene reconstruction, the performance of time efficiency tends to saturate, hence the advantage of using more robots cannot be fully exposed. To tackle this issue, we evaluate the generalization ability on the large scenes ($\geq 50m^2$) in the Matterport3D dataset. The results are shown in Table \ref{table:number_robots}. Our algorithm is able to achieve very similar performance compared to its upper bound even when it generalizes to the 9-robot scenario. It validates the exceptional generalization ability of our algorithm again.

\setlength{\tabcolsep}{8pt}
\begin{table}[t]
\centering
    \small
    \begin{tabular}{c|cc}
        \toprule
        {NeuralCoMapping (Ours)} & 
        {Cov. (\%)} & 
        {Time (\#steps)}\\
        \midrule
        single training scene & 97.1 & 691.3 \\
        nine training scenes & 96.3 & 661.7 \\
        \bottomrule
    \end{tabular}
    \vspace{2mm}
    \caption{Experiments of training with a single scene or nine scenes for our algorithm. The results are reported on the Matterport3D dataset. 
    }
\label{table:single_scene}
\vspace{-3mm}
\end{table}

\setlength{\tabcolsep}{20pt}
\begin{table*}[htb]
\centering
    \small
    \begin{tabular}{c|cc|cc}
        \toprule
        & \multicolumn{2}{c|}{Train with 3 robots} & \multicolumn{2}{c}{Upper bound} \\
        \midrule
        Test number of robots & {Cov. (\%)} & {Time (\#steps)} & {Cov. (\%)} & {Time (\#steps)} \\
        \midrule
        2 robots & 97.1 & 1293.8 & 96.6 & 1276.5\\
        4 robots & 98.4 & 798.7 & 98.4 & 776.3\\
        5 robots & 98.1 & 728.7 & 98.2 & 693.5\\
        7 robots & 96.9 & 694.0 & 98.4 & 662.3\\
        9 robots & 98.6 & 589.7 & 98.5 & 580.8\\
        \bottomrule
    \end{tabular}
    \vspace{2mm}
        \caption{Generalization to more robots on the Matterport3D dataset. Our algorithm is trained with 3 robots, and evaluated with 2, 4, 5, 7 and 9 robots separately. The upper bound performance is computed by training and evaluated on the same number of robots.}
\label{table:number_robots}
\vspace{-3mm}
\end{table*}

In this work, we focus on active mapping of indoor scenes from the Gibson and Matterport3D datasets, where usually less than 10 robots are sufficient. We further evaluate 100 robots in a scene of $633.6m^2$, and observe consistently better results of our algorithm (450 steps) than the best competitor CoScan (525 steps).

{\small
\bibliographystyle{ieee_fullname}
\bibliography{egbib}

\begin{thebibliography}{10}\itemsep=-1pt

\bibitem{bai2016information}
Shi Bai, Jinkun Wang, Fanfei Chen, and Brendan Englot.
\newblock Information-theoretic exploration with bayesian optimization.
\newblock In {\em 2016 IEEE/RSJ International Conference on Intelligent Robots
  and Systems (IROS)}, pages 1816--1822. IEEE, 2016.

\bibitem{bhattacharya2014multi}
Subhrajit Bhattacharya, Robert Ghrist, and Vijay Kumar.
\newblock Multi-robot coverage and exploration on riemannian manifolds with
  boundaries.
\newblock {\em The International Journal of Robotics Research}, 33(1):113--137,
  2014.

\bibitem{bourgault2002information}
Frederic Bourgault, Alexei~A Makarenko, Stefan~B Williams, Ben Grocholsky, and
  Hugh~F Durrant-Whyte.
\newblock Information based adaptive robotic exploration.
\newblock In {\em IEEE/RSJ international conference on intelligent robots and
  systems}, volume~1, pages 540--545. IEEE, 2002.

\bibitem{burgard2005coordinated}
Wolfram Burgard, Mark Moors, Cyrill Stachniss, and Frank~E Schneider.
\newblock Coordinated multi-robot exploration.
\newblock {\em IEEE Transactions on robotics}, 21(3):376--386, 2005.

\bibitem{caetano2009learning}
Tib{\'e}rio~S Caetano, Julian~J McAuley, Li Cheng, Quoc~V Le, and Alex~J Smola.
\newblock Learning graph matching.
\newblock {\em IEEE transactions on pattern analysis and machine intelligence},
  31(6):1048--1058, 2009.

\bibitem{cao2021tare}
Chao Cao, Hongbiao Zhu, Howie Choset, and Ji Zhang.
\newblock Tare: A hierarchical framework for efficiently exploring complex 3d
  environments.
\newblock In {\em Robotics: Science and Systems Conference (RSS), Virtual},
  2021.

\bibitem{chang2017matterport3d}
Angel Chang, Angela Dai, Thomas Funkhouser, Maciej Halber, Matthias Niessner,
  Manolis Savva, Shuran Song, Andy Zeng, and Yinda Zhang.
\newblock Matterport3d: Learning from rgb-d data in indoor environments.
\newblock {\em International Conference on 3D Vision (3DV)}, 2017.

\bibitem{chaplot2020learning}
Devendra~Singh Chaplot, Dhiraj Gandhi, Saurabh Gupta, Abhinav Gupta, and Ruslan
  Salakhutdinov.
\newblock Learning to explore using active neural slam.
\newblock In {\em International Conference on Learning Representations (ICLR)},
  2020.

\bibitem{chaplot2020object}
Devendra~Singh Chaplot, Dhiraj~Prakashchand Gandhi, Abhinav Gupta, and Russ~R
  Salakhutdinov.
\newblock Object goal navigation using goal-oriented semantic exploration.
\newblock {\em Advances in Neural Information Processing Systems}, 33, 2020.

\bibitem{chaplot2020neural}
Devendra~Singh Chaplot, Ruslan Salakhutdinov, Abhinav Gupta, and Saurabh Gupta.
\newblock Neural topological slam for visual navigation.
\newblock In {\em Proceedings of the IEEE/CVF Conference on Computer Vision and
  Pattern Recognition}, pages 12875--12884, 2020.

\bibitem{chen2013scalable}
Jiawen Chen, Dennis Bautembach, and Shahram Izadi.
\newblock Scalable real-time volumetric surface reconstruction.
\newblock {\em ACM Transactions on Graphics (ToG)}, 32(4):1--16, 2013.

\bibitem{chen2021topological}
Kevin Chen, Junshen~K Chen, Jo Chuang, Marynel V{\'a}zquez, and Silvio
  Savarese.
\newblock Topological planning with transformers for vision-and-language
  navigation.
\newblock In {\em Proceedings of the IEEE/CVF Conference on Computer Vision and
  Pattern Recognition}, pages 11276--11286, 2021.

\bibitem{chen2019learning}
Tao Chen, Saurabh Gupta, and Abhinav Gupta.
\newblock Learning exploration policies for navigation.
\newblock In {\em International Conference on Learning Representations}, 2019.

\bibitem{corah2017efficient}
Micah Corah and Nathan Michael.
\newblock Efficient online multi-robot exploration via distributed sequential
  greedy assignment.
\newblock In {\em Robotics: Science and Systems}, volume~13, 2017.

\bibitem{dai2017bundlefusion}
Angela Dai, Matthias Nie{\ss}ner, Michael Zoll{\"o}fer, Shahram Izadi, and
  Christian Theobalt.
\newblock Bundlefusion: Real-time globally consistent 3d reconstruction using
  on-the-fly surface re-integration.
\newblock {\em ACM Transactions on Graphics 2017 (TOG)}, 2017.

\bibitem{dong2019multi}
Siyan Dong, Kai Xu, Qiang Zhou, Andrea Tagliasacchi, Shiqing Xin, Matthias
  Nie{\ss}ner, and Baoquan Chen.
\newblock Multi-robot collaborative dense scene reconstruction.
\newblock {\em ACM Transactions on Graphics (TOG)}, 38(4):1--16, 2019.

\bibitem{dornhege2013frontier}
Christian Dornhege and Alexander Kleiner.
\newblock A frontier-void-based approach for autonomous exploration in 3d.
\newblock {\em Advanced Robotics}, 27(6):459--468, 2013.

\bibitem{faigl2013determination}
Jan Faigl and Miroslav Kulich.
\newblock On determination of goal candidates in frontier-based multi-robot
  exploration.
\newblock In {\em 2013 European Conference on Mobile Robots}, pages 210--215.
  IEEE, 2013.

\bibitem{faigl2012goal}
Jan Faigl, Miroslav Kulich, and Libor P{\v{r}}eu{\v{c}}il.
\newblock Goal assignment using distance cost in multi-robot exploration.
\newblock In {\em 2012 IEEE/RSJ International Conference on Intelligent Robots
  and Systems}, pages 3741--3746. IEEE, 2012.

\bibitem{hane2015obstacle}
Christian H{\"a}ne, Torsten Sattler, and Marc Pollefeys.
\newblock Obstacle detection for self-driving cars using only monocular cameras
  and wheel odometry.
\newblock In {\em 2015 IEEE/RSJ International Conference on Intelligent Robots
  and Systems (IROS)}, pages 5101--5108. IEEE, 2015.

\bibitem{holz2010evaluating}
Dirk Holz, Nicola Basilico, Francesco Amigoni, and Sven Behnke.
\newblock Evaluating the efficiency of frontier-based exploration strategies.
\newblock In {\em ISR 2010 (41st International Symposium on Robotics) and
  ROBOTIK 2010 (6th German Conference on Robotics)}, pages 1--8. VDE, 2010.

\bibitem{huang2019coloring}
Jiayi Huang, Mostofa Patwary, and Gregory Diamos.
\newblock Coloring big graphs with alphagozero.
\newblock {\em arXiv preprint arXiv:1902.10162}, 2019.

\bibitem{izadi2011kinectfusion}
Shahram Izadi, David Kim, Otmar Hilliges, David Molyneaux, Richard Newcombe,
  Pushmeet Kohli, Jamie Shotton, Steve Hodges, Dustin Freeman, Andrew Davison,
  et~al.
\newblock Kinectfusion: real-time 3d reconstruction and interaction using a
  moving depth camera.
\newblock In {\em Proceedings of the 24th annual ACM symposium on User
  interface software and technology}, pages 559--568, 2011.

\bibitem{kim2015active}
Ayoung Kim and Ryan~M Eustice.
\newblock Active visual slam for robotic area coverage: Theory and experiment.
\newblock {\em The International Journal of Robotics Research},
  34(4-5):457--475, 2015.

\bibitem{kool2018attention}
Wouter Kool, Herke Van~Hoof, and Max Welling.
\newblock Attention, learn to solve routing problems!
\newblock {\em arXiv preprint arXiv:1803.08475}, 2018.

\bibitem{kulich2011distance}
Miroslav Kulich, Jan Faigl, and Libor P{\v{r}}eu{\v{c}}il.
\newblock On distance utility in the exploration task.
\newblock In {\em 2011 IEEE International Conference on Robotics and
  Automation}, pages 4455--4460. IEEE, 2011.

\bibitem{li2021igibson}
Chengshu Li, Fei Xia, Roberto Martín-Martín, Michael Lingelbach, Sanjana
  Srivastava, Bokui Shen, Kent Vainio, Cem Gokmen, Gokul Dharan, Tanish Jain,
  Andrey Kurenkov, Karen Liu, Hyowon Gweon, Jiajun Wu, Li Fei-Fei, and Silvio
  Savarese.
\newblock igibson 2.0: Object-centric simulation for robot learning of everyday
  household tasks.
\newblock In {\em Conference on Robot Learning (CoRL)}, 2021.

\bibitem{munkres1957algorithms}
James Munkres.
\newblock Algorithms for the assignment and transportation problems.
\newblock {\em Journal of the society for industrial and applied mathematics},
  5(1):32--38, 1957.

\bibitem{newcombe2011kinectfusion}
Richard~A Newcombe, Shahram Izadi, Otmar Hilliges, David Molyneaux, David Kim,
  Andrew~J Davison, Pushmeet Kohi, Jamie Shotton, Steve Hodges, and Andrew
  Fitzgibbon.
\newblock Kinectfusion: Real-time dense surface mapping and tracking.
\newblock In {\em 2011 10th IEEE international symposium on mixed and augmented
  reality}, pages 127--136. IEEE, 2011.

\bibitem{nieto2014coordination}
Carlos Nieto-Granda, John~G Rogers~III, and Henrik~I Christensen.
\newblock Coordination strategies for multi-robot exploration and mapping.
\newblock {\em The International Journal of Robotics Research}, 33(4):519--533,
  2014.

\bibitem{ramakrishnan2020occupancy}
Santhosh~K Ramakrishnan, Ziad Al-Halah, and Kristen Grauman.
\newblock Occupancy anticipation for efficient exploration and navigation.
\newblock In {\em European Conference on Computer Vision}, pages 400--418.
  Springer, 2020.

\bibitem{rodriguez2018importance}
Mar{\'\i}a~L Rodr{\'\i}guez-Ar{\'e}valo, Jos{\'e} Neira, and Jos{\'e}~A
  Castellanos.
\newblock On the importance of uncertainty representation in active slam.
\newblock {\em IEEE Transactions on Robotics}, 34(3):829--834, 2018.

\bibitem{schulman2017proximal}
John Schulman, Filip Wolski, Prafulla Dhariwal, Alec Radford, and Oleg Klimov.
\newblock Proximal policy optimization algorithms.
\newblock {\em arXiv preprint arXiv:1707.06347}, 2017.

\bibitem{sethian1996fast}
James~A Sethian.
\newblock A fast marching level set method for monotonically advancing fronts.
\newblock {\em Proceedings of the National Academy of Sciences},
  93(4):1591--1595, 1996.

\bibitem{shen2021igibson}
Bokui Shen, Fei Xia, Chengshu Li, Roberto Martín-Martín, Linxi Fan, Guanzhi
  Wang, Claudia Pérez-D'Arpino, Shyamal Buch, Sanjana Srivastava, Lyne~P.
  Tchapmi, Micael~E. Tchapmi, Kent Vainio, Josiah Wong, Li Fei-Fei, and Silvio
  Savarese.
\newblock igibson 1.0: a simulation environment for interactive tasks in large
  realistic scenes.
\newblock In {\em IEEE/RSJ International Conference on Intelligent Robots and
  Systems (IROS)}, 2021.

\bibitem{sinkhorn1967concerning}
Richard Sinkhorn and Paul Knopp.
\newblock Concerning nonnegative matrices and doubly stochastic matrices.
\newblock {\em Pacific Journal of Mathematics}, 21(2):343--348, 1967.

\bibitem{stachniss2005information}
Cyrill Stachniss, Giorgio Grisetti, and Wolfram Burgard.
\newblock Information gain-based exploration using rao-blackwellized particle
  filters.
\newblock In {\em Robotics: Science and systems}, volume~2, pages 65--72, 2005.

\bibitem{sykora2020multi}
Quinlan Sykora, Mengye Ren, and Raquel Urtasun.
\newblock Multi-agent routing value iteration network.
\newblock In {\em International Conference on Machine Learning}, pages
  9300--9310. PMLR, 2020.

\bibitem{tabib2016computationally}
Wennie Tabib, Micah Corah, Nathan Michael, and Red Whittaker.
\newblock Computationally efficient information-theoretic exploration of pits
  and caves.
\newblock In {\em 2016 IEEE/RSJ International Conference on Intelligent Robots
  and Systems (IROS)}, pages 3722--3727. IEEE, 2016.

\bibitem{vaswani2017attention}
Ashish Vaswani, Noam Shazeer, Niki Parmar, Jakob Uszkoreit, Llion Jones,
  Aidan~N Gomez, {\L}ukasz Kaiser, and Illia Polosukhin.
\newblock Attention is all you need.
\newblock In {\em Advances in neural information processing systems}, pages
  5998--6008, 2017.

\bibitem{visser2013discussion}
Arnoud Visser, Julian De~Hoog, Adrian Jim{\'e}nez-Gonz{\'a}lez, and
  J-R~Martinez de Dios.
\newblock Discussion of multi-robot exploration in communication-limited
  environments.
\newblock In {\em Workshop” Towards Fully Decentralized Multi-Robot Systems:
  Hardware, Software and Integration” at the ICRA Conference}. Citeseer,
  2013.

\bibitem{wang2019learning}
Runzhong Wang, Junchi Yan, and Xiaokang Yang.
\newblock Learning combinatorial embedding networks for deep graph matching.
\newblock In {\em Proceedings of the IEEE/CVF International Conference on
  Computer Vision}, pages 3056--3065, 2019.

\bibitem{werger2000broadcast}
Barry~Brian Werger and Maja~J Matari{\'c}.
\newblock Broadcast of local eligibility for multi-target observation.
\newblock In {\em Distributed Autonomous Robotic Systems 4}, pages 347--356.
  Springer, 2000.

\bibitem{wu2020comprehensive}
Zonghan Wu, Shirui Pan, Fengwen Chen, Guodong Long, Chengqi Zhang, and S~Yu
  Philip.
\newblock A comprehensive survey on graph neural networks.
\newblock {\em IEEE transactions on neural networks and learning systems},
  32(1):4--24, 2020.

\bibitem{xia2018gibson}
Fei Xia, Amir~R Zamir, Zhiyang He, Alexander Sax, Jitendra Malik, and Silvio
  Savarese.
\newblock Gibson env: Real-world perception for embodied agents.
\newblock In {\em Proceedings of the IEEE Conference on Computer Vision and
  Pattern Recognition}, pages 9068--9079, 2018.

\bibitem{yamauchi1997frontier}
Brian Yamauchi.
\newblock A frontier-based approach for autonomous exploration.
\newblock In {\em Proceedings 1997 IEEE International Symposium on
  Computational Intelligence in Robotics and Automation CIRA'97.'Towards New
  Computational Principles for Robotics and Automation'}, pages 146--151. IEEE,
  1997.

\bibitem{ye2021auxiliary}
Joel Ye, Dhruv Batra, Abhishek Das, and Erik Wijmans.
\newblock Auxiliary tasks and exploration enable objectgoal navigation.
\newblock In {\em Proceedings of the IEEE/CVF International Conference on
  Computer Vision}, pages 16117--16126, 2021.

\bibitem{yu2021learning}
Chao Yu, Xinyi Yang, Jiaxuan Gao, Huazhong Yang, Yu Wang, and Yi Wu.
\newblock Learning efficient multi-agent cooperative visual exploration.
\newblock {\em arXiv preprint arXiv:2110.05734}, 2021.

\bibitem{zhang2020multiplex}
Weifeng Zhang, Jingwen Mao, Yi Cao, and Congfu Xu.
\newblock Multiplex graph neural networks for multi-behavior recommendation.
\newblock In {\em Proceedings of the 29th ACM International Conference on
  Information \& Knowledge Management}, pages 2313--2316, 2020.

\end{thebibliography}
}

\end{document}